\documentclass[times,final]{elsarticle}

\usepackage{dib}
\usepackage{framed,multirow}

\usepackage{amssymb}
\usepackage{latexsym}
\usepackage{tikz} 
\usepackage{pgf-pie}

\usepackage{url}
\usepackage{xcolor}
\definecolor{newcolor}{rgb}{.8,.349,.1}

\usepackage{longtable}
\usepackage[colorlinks]{hyperref}

\journal{Data in Brief}

\begin{document}

\verso{F. Gasparini \textit{et al.}}

\begin{frontmatter}

\dochead{Data Article}
\title{Benchmark dataset of memes with text transcriptions for automatic detection of multi-modal misogynistic content}



\author{Francesca \snm{Gasparini}}
\ead{francesca.gasparini@unimib.it}
\author{Giulia \snm{Rizzi}}
\ead{g.rizzi10@campus.unimib.it}
\author{Aurora \snm{Saibene}}
\ead{a.saibene2@campus.unimib.it}
\author{Elisabetta \snm{Fersini}\corref{cor1}}
\ead{elisabetta.fersini@unimib.it}
\cortext[cor1]{Corresponding author: 
  Tel.: +39-02-6448-7896.}

\address{Department of Informatics, Systems and Communication,  University of Milano-Bicocca, Italy}


\begin{abstract}
In this paper we present a benchmark dataset generated as part of a project for automatic identification of misogyny within online content, which focuses in particular on memes. 
The benchmark here described is composed of 800 memes collected from the most popular social media platforms, such as Facebook,
Twitter, Instagram and Reddit, and consulting websites dedicated to collection and creation of memes. 
To gather misogynistic memes, 
specific keywords that refer to misogynistic content have been considered as search criterion, considering different manifestations of hatred against women, such as body shaming, stereotyping, objectification and violence. In parallel, memes with no misogynist content have been manually downloaded from the same web sources. 
Among all the collected memes, three domain experts have selected a dataset of 800 memes equally balanced between misogynistic and non-misogynistic ones.
This dataset has been validated through a  crowdsourcing platform, involving 60 subjects for the labelling
process, in order to collect three evaluations for each instance. 
Two further binary labels have been collected from both the experts and the crowdsourcing platform, for memes evaluated as misogynistic, concerning aggressiveness and irony.
Finally for each meme, the text has been manually transcribed. 
The dataset provided is thus composed of the 800 memes, the labels given by the experts and those obtained by the crowdsourcing validation, and the transcribed texts. 
This data can be used to approach the problem of automatic detection of misogynistic content on the Web relying on both textual and visual cues, facing  phenomenons that are growing every day such as cybersexism and technology-facilitated violence.
\end{abstract}


\end{frontmatter}

{\fontsize{7.5pt}{9pt}\selectfont
\noindent\textbf{Specification Table}

\begin{longtable}{|p{33mm}|p{94mm}|}
\hline
\endhead
\hline
\endfoot
Subject                & Artificial Intelligence\\
\hline                         
Specific subject area  & Automatic detection of multi-modal content\\
\hline
Type of data           & Tables (.csv)\newline
                         Images (.jpeg)\newline
 \\                                     
How data were acquired & Images of memes were downloaded from the Web  and saved in high quality JPG format with maximum dimension resized to 640 pixels in order to have a dataset of normalized dimensions; labelled data were collected from domain experts and through a crowdsourcing platform; text data were manually transcribed.   \\
\hline                         
Data format            &                       Raw data\newline
                         Analyzed data (labeled and transcribed) 
\\
\hline                         
Parameters for         
data\newline 
collection             & Memes were downloaded from popular social media platforms and websites dedicated to the collection and  creation  of  memes, using keywords related to misogyny. The memes were subsequently  manually selected. Each meme was evaluated by three domain experts and by three subjects through a crowdsourcing platform.\\  

\hline
Description of          
data\newline 
collection   & The 800 memes of the dataset were selected from a bigger collection, by a pool of three experts of the domain in order to be balanced  between  misogynistic  and  non-misogynistic ones. The dataset has been validated by using the Figure Eight crowdsourcing platform (now called Appen, https://appen.com/), involving 60 subjects, and the text of each meme has been manually transcribed.\\
\hline                         
Data source location   & 
Raw data as well as crowdsourcing labels were collected from the Web\newline
Data were processed at: \newline
                         Institution:University of Milano-Bicocca\newline
                         City: Milan\newline
                         Country: Italy
                                               \\
\hline                         
\hypertarget{target1}
{Data accessibility}   & In a public repository:\newline
                         Repository name: github \newline
                         Data identification name: Misogynistic-MEME\newline
                         Direct URL to data: \url{https://github.com/MIND-Lab/MEME} \newline
                                                 Instructions for accessing these data:\newline
                        Data are password protected. Password will be provided after fulfilling a copyright notice.\\                         
\hline                         
Related                 
research\newline
article                & E. Fersini, F. Gasparini, S. Corchs, Detecting sexist MEME on the Web: A study on textual and visual cues. In: 2019 8th International Conference on Affective Computing and Intelligent Interaction Workshops and Demos (ACIIW). IEEE, 2019. p. 226-231.
\end{longtable}
}
\newpage
\section*{Value of the Data}


\begin{itemize}
\itemsep=0pt
\parsep=0pt
\item The role of women within the society has increased in importance and the way we approach and refer to them is crucial. Misogyny is a form of discrimination towards women and has been spreading exponentially through
the Web. A popular communication tool in social media platforms are memes which are typically composed of pictorial 
and textual components, and that are often used to convey  misogynistic messages. Automatic detection of misogynistic content is becoming  fundamental to counteract online discrimination, cybersexism and violence. 
As misogyny is conveyed by both textual and visual media, a dataset of multimodal content, such as a dataset of memes, is mandatory to build efficient machine learning techniques that promptly intercept these offensive messages online. 
\item The presented dataset will be particularly useful for: i) researchers from Natural Language Processing and Computer Vision communities that deal with social media data analysis and that are interested in developing machine learning models that combine textual and visual cues; ii) companies that develop artificial intelligent strategies to control social media activities; and iii) social science researchers that study gender discrimination especially on the Web.
\item  This dataset presents a data structure that can be adopted for the collection of further data on the topic to develop more robust machine learning techniques. In fact the automatic detection of misogynistic content is particularly challenging especially considering that: (i) misogynistic and non misogynistic memes can share the same visual content but a different text, and (ii) misogyny can be expressed by text, image or by their combination.  
This dataset can also be adopted to increase the quality of the labelling task, as it can be used as a gold standard to check the reliability of annotators on  crowdsourcing platforms. 
\item This dataset will pave the way not only to the development of machine learning techniques able to promptly intercept on offensive content  against women the Web, but also to stimulate other researchers in devoting their attention to this phenomenon, increasing the awareness and sensitivity to misogyny as well as to other forms of discrimination.  
\item This dataset provides  the labels from both domain experts and subjects recruited within the population. The comparison between the two distributions is also significant, not only with respect to misogyny by itself, but also with respect to the perception of  aggressiveness and irony. 
\item Up to our knowledge, this is the first dataset of memes that face the misogynist phenomenon. A related dataset is the one proposed by Facebook AI for the Hateful Meme (HM) Challenge \cite{kiela2021hateful}. The main difference between the HM dataset and the proposed one is that  our memes have been collected from social media platforms, describing a real scenario, while the others could refer to synthetic memes automatically generated as benign confounders of similar hateful memes. Moreover, while the HM dataset is generally devoted to general hateful and non-hateful memes, without focusing on any specific types and targets of hate, our dataset is centered on women as target. 

\end{itemize}

\section*{Data Description}

In the latest years, misogyny has found in the Web a new and powerful way of diffusion together with the new phenomenon of cybersexism, where women are often victim of offensive messages and, in the most serious cases, of abuse and threats.
Online platform providers have introduced policies to prevent offensive content. However, due to the speed of dissemination of messages in social media,  systems able to automatically filter offensive content are urgently needed \cite{anzovino,gasparini}. 
While new opportunities for females have been opened on the Web, systematic inequality and  discrimination offline is replicated in online spaces 
in the form of offensive contents against them \cite{frenda,plaza}. \\
Memes are especially popular communication means for social media users, being able to efficaciously convey funny and/or ironic jokes \cite{memeDef}. A meme is defined as an image composed of a pictorial information on which a text is superimposed a posteriori by a human \cite{memeDef}. Given the characterization of memes, they have been progressively used to convey hate \cite{farrell} and, in this specific dataset, we are interested in memes circulating in the Web and targeting women with sexist and aggressive messages \cite{paciello2021online, franks2011unwilling}.\\
%
In order to develop efficient machine learning techniques able to automatically detect multi-modal misogynistic messages online, we here present a dataset composed of: 
 \begin{itemize}
     \item \textbf{800 memes}, saved as jpeg images, resized to have the greatest dimension equal to 640 pixels. These memes are saved with a progressive unique ID.
     \item A \textbf{table} saved as a .csv file, where all the data collected are reported, according to the following structure: 
     \begin{itemize}
        \item   \textit{memeID}: unique identifier associated to the meme;
        \item  \textit{text}: transcription of the text reported in the meme;
        \item    \textit{misogynisticDE}: Boolean attribute related to the presence of misogynistic content as reported by the  Domain Experts (DE);
        \item \textit{aggressiveDE}: Boolean attribute; in case of a misogynist meme it represents the presence of aggressiveness, as reported by the DE;
        \item \textit{ironicDE}: Boolean attribute; in case of a misogynist meme it represents the presence of irony, as reported by the DE;
        \item \textit{misogynisticCS}: Boolean attribute related to the presence of misogynistic content, as reported by the annotators of the CrowdSourcing  platform (CS);
        \item \textit{aggressiveCS}: Boolean attribute; in case of misogynist meme it represents the presence of aggressiveness, as reported by the CS;
        \item  \textit{ironicCS}: Boolean attribute; in case of misogynist meme it represents the presence of irony, as reported by the CS.
        \item \textit{confidence\_M\_CS}: agreement on the misogynist attribute among the CS;
        \item \textit{confidence\_A\_CS}: agreement on the aggressiveness attribute among the CS;
        \item \textit{confidence\_I\_CS}: agreement on the misogynist irony among the CS;

    \end{itemize}
 \end{itemize}
 
 \noindent We underline that the labels of the 3 experts have an agreement of 100\% for all the 800 memes, as it was the criterion to select them among all the downloaded memes. 

\begin{figure}
    \centering
    \includegraphics[width=0.9 \columnwidth]{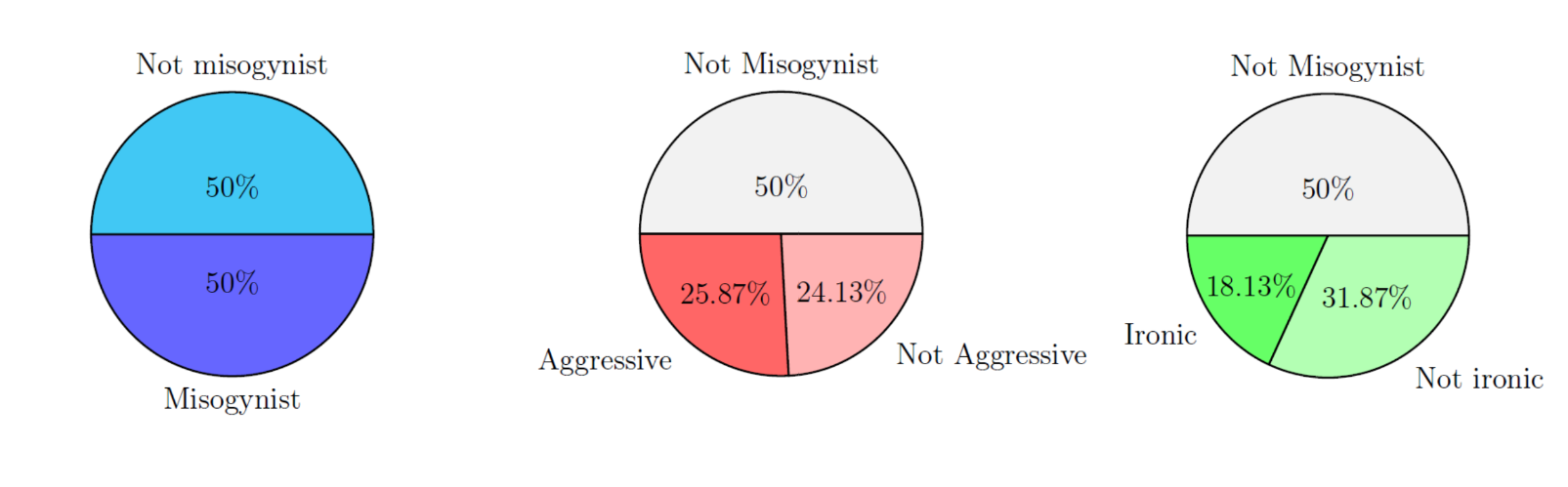}
    \caption{Distribution of the three labels given by the domain experts (DE)}
    \label{fig:pieED}
\end{figure}

\paragraph{Data Distribution}
The distribution of the three labels (misogyny, aggressiveness and irony) given by the domain experts are reported in Figure \ref{fig:pieED}. As the dataset of 800 memes was selected starting from the misogynistic DE labels, the first pie chart on the left confirms that the dataset is equally distributed within the two classes with 400 memes each. 
The second and third pie charts reported for the memes labelled as misogynistic the percentage of them considered respectively aggressive and ironic. 

\begin{figure}
    \centering
    \includegraphics[width=0.9 \columnwidth]{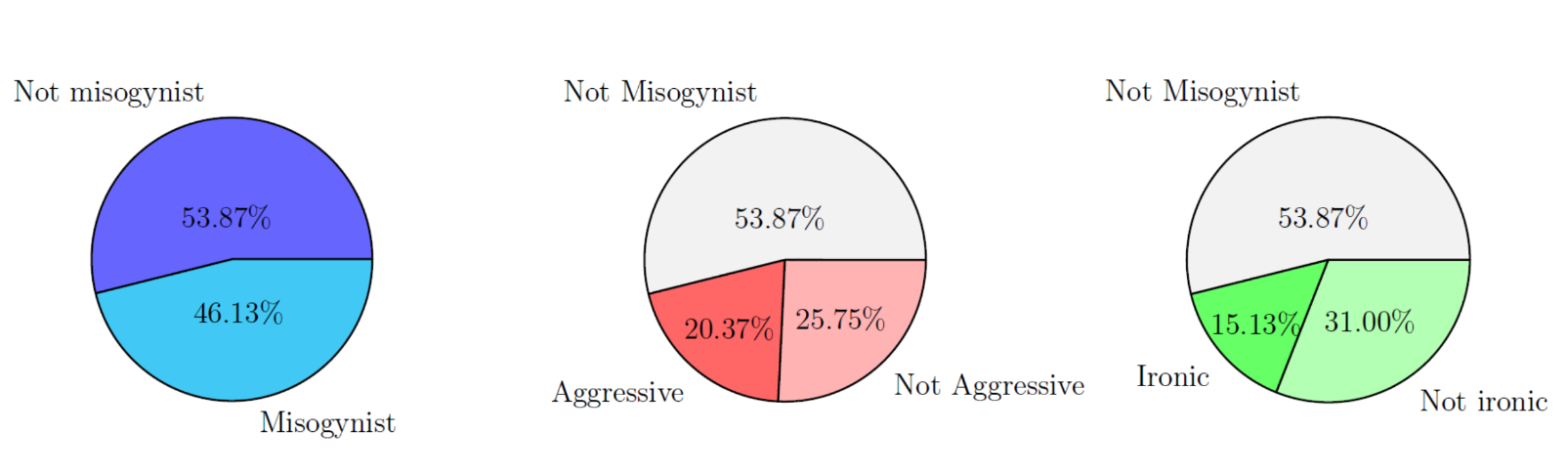}
    \caption{Distribution of the three labels given by the crowdsourcing annotators (CS)}
    \label{fig:pieCS}
\end{figure}

In Figure \ref{fig:pieCS}, the corresponding label distributions given by the CS annotators are reported.
The first pie on the left shows that in this case the memes labelled as misogynistic are less than in the case of the DE evaluation. 

Given the labels provided by DE and CS, 59 memes have been annotated differently with respect to the misogynistic label. Among them, 76.28\% were considered misogynistic by the experts but not by the annotators, while only 23.72 \% of them were considered misogynistic only by the annotators. 
The complexity of the annotation process is also reflected by the agreements of the CS labellers, which have been reported in the corresponding columns in the Annotation Data sheet.

\section*{Experimental Design, Materials and Methods}

\subsection*{Meme collection process}

\noindent The most popular social media platforms, i.e. Facebook, Twitter, Instagram and Reddit, have been considered in the data collection phase. Memes that convey potential misogynistic content have been collected in the October-November 2018 period through the following operations \cite{ACIIW}:
\begin{itemize}
    \item Searching for threads or conversations dedicated and written by anti-women/feminists supporters, such as the Men Going Their Own Way (MGTOW) website and the related thread on Reddit;
    \item Exploring discussions on sexism in political or social events;
    \item Browsing hastags such as \#girl, \#girlfriend, \#women.
\end{itemize}
Subsequently, the dataset has been enlarged by collecting memes from websites dedicated to meme creation and/or collection, as follows:
\begin{itemize}
    \item Browsing hashtags such as \#girl, \#girlfriend, \#women, \#feminist;
    \item Consulting collections on all the variations of famous memes involving female characters.
\end{itemize}

\noindent In parallel, memes with non-misogynistic contents have been manually downloaded from the same web sources and by adopting the same keywords, for a non trivial collection of the memes.

\subsection*{Expert labelling and dataset definition}

\noindent Three domain experts have evaluated all the collected memes, labelling them as misogynist or non-misogynist. 
The final dataset is composed of 800 memes, selected among those with an agreement of 100\%, in order to have a perfect balanced dataset with respect to the two classes.

\noindent The experts have also annotated the misogynistic memes with respect to aggressiveness and irony. This phase provided the three corresponding boolean labels reported in the data sheet with DE suffix.   

\subsection*{Dataset Annotation through the crowdsourcing platform}

\noindent The 800 memes selected were labelled adopting a crowdsourcing platform (Figure Eight in 2018, now called Appen, https://appen.com/).\\
A controlled labelling experiment was chosen to provide judgments from trusted and reliable participants with equally distributed age (between 20 and 50 years old) and gender. All the data were anonymized.
%

\noindent The annotation task was designed as follows: 

\begin{itemize}
    \item the order of the memes in the experiment is randomized to avoid bias;
    \item the maximum number of judgments that any contributor can provide is limited to 40 memes, to limit fatigue and consequent unreliable annotations; 
    \item any annotator can leave whenever he/she wants;
    \item for each annotator, the task expires after an hour and a half, regardless of the number of evaluated memes, in order to limit external stimuli; 
    \item each annotation page shows only one meme at a time, to not influence or bias the participant by seeing other meme contemporaneously;
    \item each meme is evaluated by three different subjects.
\end{itemize}

\begin{figure}
    \centering
    \includegraphics[width=0.4\columnwidth]{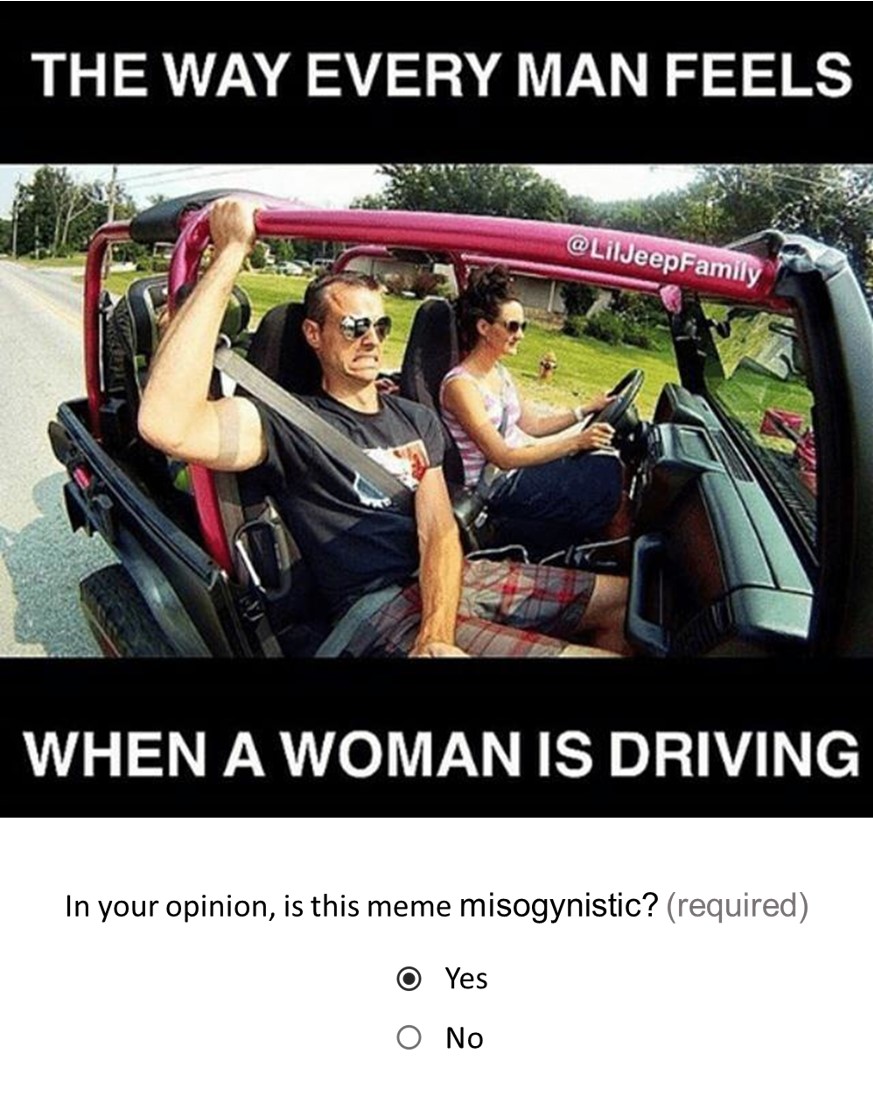}
    \caption{The first question in the crowdsourcing annotation task}
    \label{fig:sexist}
\end{figure}

\noindent For each meme, the question 
\textbf{In your opinion, is this meme misogynistic?} is proposed to the participant, as depicted in 
Figure \ref{fig:sexist}. 

\noindent Then, only in the case of a meme evaluated as misogynistic, the following two questions were proposed: 

\begin{enumerate}
    \item In your opinion, is this meme ironic?
    \item In your opinion, is this meme aggressive?
\end{enumerate}

\noindent Definitions and guidelines were not provided to the CS annotators, wanting to collect their perception of misogyny, irony and  aggressiveness present in the meme, without influencing their decision.
This crowdsourcing annotation  provided the three corresponding boolean labels reported in the data sheet with CS suffix, together with the corresponding confidence level (in terms of percentage of agreement).

\subsection*{Text transcription}
\noindent Finally, the text superimposed to the imaged have been manually transcribed for each of the 800 memes.

\section*{Ethics Statement}
\noindent An informed consent was given to all the involved subjects, indicating the presence of possible explicit contents,  explaining how to perform the tasks also reporting  them the possibility to stop the labeling activity whenever they wanted. 
Any personal data have been acquired to identify the labellers, being therefore GDPR compliant.

\section*{Acknowledgments}
\noindent We want to give our thanks especially to Silvia Corchs for her valuable scientific support in facing the problem of misogyny detection, and to  Monica Mantovani and Gaia Campisi, for their supporting work during the collection and validation of the dataset.

\section*{Declaration of Competing Interest}
\noindent
The authors declare that they have no known competing
financial interests or personal relationships which have, or could be
perceived to have, influenced the work reported in this article. 


\bibliographystyle{model1-num-names}
\bibliography{GoldStandard2}


\end{document}